\pdfoutput=1

\documentclass[11pt]{article}

\usepackage[table]{xcolor}
\usepackage[preprint]{acl}

\usepackage{times}
\usepackage{latexsym}

\usepackage[T1]{fontenc}

\usepackage[utf8]{inputenc}
\usepackage{microtype}
\usepackage{inconsolata}
\usepackage{graphicx}
\usepackage{booktabs}
\usepackage{amsmath}
\usepackage{amssymb}
\usepackage{subfigure}
\usepackage{multirow}

\title{Cross-lingual Transfer of Reward Models in Multilingual Alignment}

\author{
    {Jiwoo Hong}\thanks{Equal Contribution}$^\dagger$\quad {Noah Lee}\footnotemark[1]$^\dagger$ \quad  {Rodrigo Martínez-Castaño}$^{\S}$\vspace{0.05in} \\ {\textbf{César Rodríguez}}$^{\S}$ \quad  {\textbf{James Thorne}}$^\dagger$
    \vspace{0.15in}
    \\
    \normalsize $^\dagger$\textsc{KAIST AI} \quad $^{\S}$\textsc{IQ.Wiki} 
    \vspace{0.1in}
    \\
    {$^{\dagger}$\url{{jiwoo\_hong, noah.lee, thorne}@kaist.ac.kr}} \\
    {$^{\S}$\url{{rodrigo, cesar}@iq.wiki}}
}

\begin{document}
\maketitle
\begin{abstract}

Reinforcement learning with human feedback (RLHF) is shown to largely benefit from precise reward models (RMs). However, recent studies in reward modeling schemes are skewed towards English, limiting the applicability of RLHF in multilingual alignments. In this work, we investigate the cross-lingual transfer of RMs trained in diverse languages, primarily from English. Our experimental results demonstrate the strong cross-lingual transfer of English RMs, exceeding target language RMs by 3$\sim$4\% average increase in \texttt{Mutlilingual RewardBench}. Furthermore, we analyze the cross-lingual transfer of RMs through the representation shifts. Finally, we perform multilingual alignment to exemplify how cross-lingual transfer in RM propagates to enhanced multilingual instruction-following capability, along with extensive analyses on off-the-shelf RMs. We release the code\footnote{Code - \href{https://github.com/IQ-KAIST/rm-lingual-transfer}{IQ-KAIST/rm-lingual-transfer}}, model and data\footnote{Data \& Models -  \href{https://huggingface.co/collections/iqwiki-kor/}{HF Collection}}.

\end{abstract}

\section{Introduction}

Recent advances in reinforcement learning with human feedback (RLHF) as a large language model (LLM) post-training technique \citep{NIPS2017_d5e2c0ad, ziegler2020finetuninglanguagemodelshuman} highlight the importance of having high-quality data \citep{wang-etal-2024-helpsteer, dubey2024llama3herdmodels} and reward model (RM) \citep{pmlr-v162-ethayarajh22a, gao2023scaling, ji2023beavertails, secrets2rm, wang2024helpsteer2opensourcedatasettraining}. Leveraging synthetic data has contributed to building stronger English RMs due to their efficiency and scalability \citep{ cui2024ultrafeedbackboostinglanguagemodels, ArmoRM, zhu2024starlingb}.

Nevertheless, adopting RMs for non-English languages is heavily understudied. While LLM-as-a-Judge can be used as a generative reward model for multilingual RLHF settings \citep{son2024llm}, generative RMs have been shown to underperform traditional RMs \citep{rewardbench, ArmoRM}. Meanwhile, \citet{wu2024reuse} empirically demonstrates the possibilities of cross-lingual transfer in RMs, but the findings were limited to simple tasks and encoder-decoder models.

In this paper, we show that RMs trained on English-only datasets (\emph{i.e.,} English RMs) display strong cross-lingual transfer when built on top of multilingual pre-trained language models (MLMs). We first demonstrate the cross-lingual transfer of English RMs by consistently outperforming target language RMs in \texttt{Multilingual RewardBench}. Then, we explain it with two reasons: \textbf{1)} English preserves representations of the initial MLMs (Section \ref{subsec:represent}), and \textbf{2)} representations of MLMs inherently have a strong understanding of languages (Section \ref{subsec:emb}), concluding that RMs should preserve representations of MLMs for generalizability. Additional analysis of off-the-shelf RMs supports our findings by both classifier and generative RMs based on MLMs having strong cross-lingual transfer. Finally, multilingual alignment experiments exhibit the propagation of strong cross-lingual transfer in English RMs to downstream usage, having an average win rate increase of 9.5\% across four non-English languages.


\begin{table*}[t!]
\centering
\begin{small}
\begin{center}
\resizebox{\textwidth}{!}{%
\begin{tabular}{c|c|ccccc|ccccc}
\toprule
                         &          & \multicolumn{5}{c|}{\textbf{\textsc{Llama-3.2-3B-IT}}}     & \multicolumn{5}{c}{\textsc{\textbf{Qwen2.5-3B-IT}}}         \\ 
\textbf{\texttt{RewardBench}}                 & \textbf{Category} & \textbf{Chat}  & \textbf{Chat(H)} & \textbf{Safety} & \textbf{Reason} & \textbf{Avg.} & \textbf{Chat}  & \textbf{Chat(H)} & \textbf{Safety} & \textbf{Reason} & \textbf{Avg.} \\ \midrule
\multirow{3}{*}{\textsc{\textbf{Spanish}}} & \textbf{Target}   & 79.1  & 67.3    & 88.0   & 65.5   & 75.0 & 80.7  & 68.2    & 84.8   & 68.2   & 75.5 \\
                         & \textbf{English}  & {}86.3  & 69.3    & 89.3   & 72.4   & 79.3 & 82.7  & 68.0    & 88.3   & 73.6   & 78.1 \\
                         & $\Delta$    &+7.2	&+2.0	&+1.3	&+6.9	&+4.3	&+2.0	&-0.2	&+3.5	&+5.4	&+2.6  \\ \midrule
\multirow{3}{*}{\textsc{\textbf{Italian}}} & \textbf{Target}   & 75.4  & 62.5    & 88.5   & 65.7   & 73.0 & 77.1  & 67.8    & 85.7   & 72.8   & 75.8 \\
                         & \textbf{English}  & 83.0  & 69.3    & 88.7   & 75.1   & 79.0 & 83.2  & 68.2    & 88.4   & 76.0   & 79.0 \\
                         & $\Delta$    &+7.6	&+6.8	&+0.2	&+9.4	&+6.0	&+6.1	&+0.4	&+2.7	&+3.2	&+3.2 \\ \midrule
\multirow{3}{*}{\textsc{\textbf{Korean}}}  & \textbf{Target}   & 69.6  & 58.8    & 80.9   & 60.1   & 67.3 & 68.4  & 63.2    & 80.9   & 61.4   & 68.5 \\
                         & \textbf{English}  & 69.8  & 59.4    & 84.3   & 73.0   & 71.6 & 70.7  & 61.6    & 85.4   & 73.6   & 72.8 \\
                         & $\Delta$    & +0.2	&+0.6	&+3.4	&+12.9	&+4.3	&+2.3	&-1.6	&+4.5	&+12.2	&+4.3  \\ \midrule
\multirow{3}{*}{\textsc{\textbf{Chinese}}} & \textbf{Target}   & 68.7  & 59.9    & 81.2   & 52.6   & 65.6 & 69.8  & 64.7    & 81.8   & 61.3   & 69.4 \\
                         & \textbf{English}  & 54.7  & 64.0    & 82.6   & 79.3   & 70.2 & 58.7  & 67.8    & 84.3   & 78.2   & 72.2 \\
                         & $\Delta$    & -14.0	&+4.1	&+1.4	&+26.7	&+4.6	&-11.1	&+3.1	&+2.5	&+16.9	&+2.8\\ \bottomrule
\end{tabular}%
}
\vspace{-0.1in}
\end{center}
\end{small}
\caption{\texttt{Multilingual RewardBench} evaluation results on the target language ("Target") and English ("English") RMs. "$\Delta$" denotes the accuracy gain of English RMs compared to the target language RMs. English RMs show higher average scores in the lingual axis than target language RMs. Also, English RMs excel target language RMs in reasoning ("Reason") tasks with diverse evaluation sub-categories.}
\label{tab:rb_full}
\end{table*}

\section{English as a \emph{Lingua Franca} in RMs}\label{sec:exp}

We empirically verify the cross-lingual transfer in reward models (RMs) trained with different languages, thereby showing that the English preference data is a \emph{lingua franca} in reward modeling.

\subsection{Background}

\paragraph{Cross-lingual transfer} 

Training multilingual language models (MLMs) at scale has shown to incur \textit{cross-lingual transfer} in both encoder-only \citep{devlin-etal-2019-bert, conneau-etal-2020-unsupervised, chi-etal-2022-xlm} and encoder-decoder \citep{xue-etal-2021-mt5} transformer architectures. Recently, studies revealed the implications of cross-lingual transfer in decoder-only models as well \citep{ustun-etal-2024-aya, wang-etal-2024-probing-emergence}; however, they were limited to generative tasks \citep{zhang-etal-2024-plug} or downstream alignment-tuning only \citep{dang2024rlhf}.

\paragraph{Reward modeling}Reward models are trained as a classifier \citep{NIPS2017_d5e2c0ad} to return a scalar value $r_\theta(\cdot)$  with the objective with the Bradley-Terry model \citep{bradley1952rank}:
\begin{equation*}
    \mathcal{L}_\text{RM} = \sigma\left(r_\theta(x, y_w) - r_\theta(x, y_l) \right),
\end{equation*}
with the prompt $x$ and corresponding preferred and dispreferred responses $y_w$ and $y_l$. While crucial in alignment-tuning \citep{rafailov2024direct, hong2024orpo, meng2024simpo}, reward modeling schemes for multilingual usage are still understudied. 
Motivated by this research opportunity, we study the cross-lingual transfer of English-focused RMs with recent autoregressive models and how it propagates to downstream multilingual alignment.

\subsection{Experimental Details}

\paragraph{Dataset} We curate a synthetic preference dataset of 86k instances\footnote{Refer to Appendix \ref{apdx:data} for detailed process.} from five representative English preference datasets: SafeRLHF \citep{safe-rlhf}, WildGuard \citep{wildguard2024}, HelpSteer2 \citep{wang2024helpsteer2opensourcedatasettraining}, Offsetbias \citep{park2024offsetbias}, and Magpie \citep{xu2024magpiealignmentdatasynthesis}. Using English data, we create four parallel machine-translated versions\footnote{Spanish (Sp), Italian (It), Korean (Ko), and Chinese (Ch)}, utilizing X-ALMA \citep{xu2024xalmaplugplay}.

\paragraph{Models} Two state-of-the-art 3B multilingual pre-trained language models are fine-tuned\footnote{Refer to Appendix \ref{apdx:config} for detailed hyperparameters.} as reward models: Llama-3.2-3B-Instruct \citep{dubey2024llama3herdmodels} and Qwen2.5-3B-Instruct \citep{qwen2}. 

\paragraph{Evaluation} We prepare four non-English \texttt{Multilingual} \texttt{RewardBench} by translating \texttt{RewardBench} \citep{rewardbench} to assess the cross-lingual transfer in RMs.

\subsection{Results and Analysis}\label{subsec:results}

\paragraph{English RMs show strongest cross-lingual transfer}Average reward model accuracy ("Avg") in Table \ref{tab:rb_full} shows that English RMs surpass target language RMs in general. Specifically, Llama-3.2-3B gained at least 4.3\%, where the cross-lingual generalizability of English RMs is more highlighted than Qwen2.5-3B, which gained at most 4.3\%. However, considering that all Qwen-based target language RMs outperform the Llama-based target language RMs, Qwen2.5-3B is shown to be a better model choice for training a language-specific RM.

\paragraph{Reasoning tasks significantly benefit from cross-lingual transfer}Generalizability of English RMs is best highlighted in the reasoning tasks ("Reason") in Table \ref{tab:rb_full}, especially in non-Latin languages. Non-Latin languages, Korean and Chinese, improved significantly in English RMs compared to target language RMs, exceeding 12\% and 27\% in Chinese, for instance.

\section{Analysis on Lingual Tranfer of MLM}\label{sec:analysis}
This section provides empirical and theoretical insights on \emph{why} English is \emph{lingua franca} in reward modeling, given a multilingual language model (MLM) using two arguments: \textbf{1)} English acts as a \emph{lingua franca} in reward modeling because it best preserves the representations of the base model, and \textbf{2)} representations in MLMs \emph{should} be preserved since they are inherently effective in language-aware encoding.

\begin{figure}[t!]
    \centering
    \subfigure[Llama-3.2-3B]{\includegraphics[width=0.49\columnwidth]{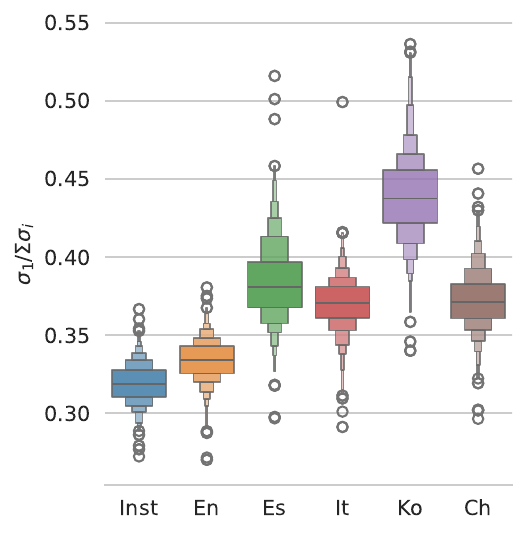}\label{subfig:emb_llama}}  
    \subfigure[Qwen2.5-3B]{\includegraphics[width=0.49\columnwidth]{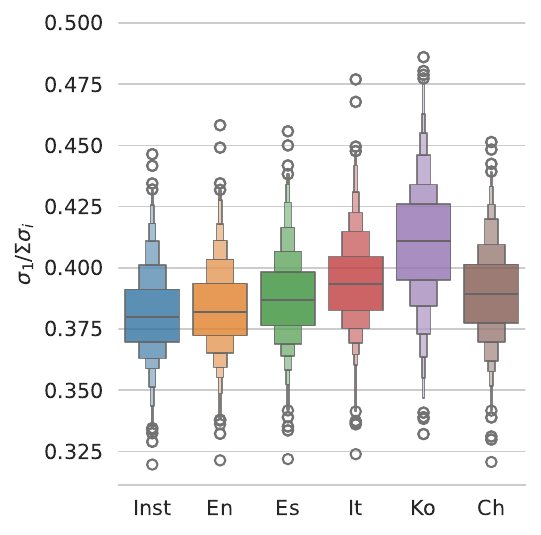}\label{subfig:emb_qwen}}  
    \vspace{-0.15in}
    \caption{Proportion of the largest singular value in the concatenated hidden states for fixed context translated in five languages with RMs trained in each language. While English ("En") best preserves the representation diversity of the base model ("Inst'), Korean ("Ko") leads to the most homogeneous representations.} 
    \label{fig:emb}
\end{figure}

\begin{figure*}[t!]
    \centering
    \subfigure[OLMo-1B]{\includegraphics[width=0.2455\textwidth]{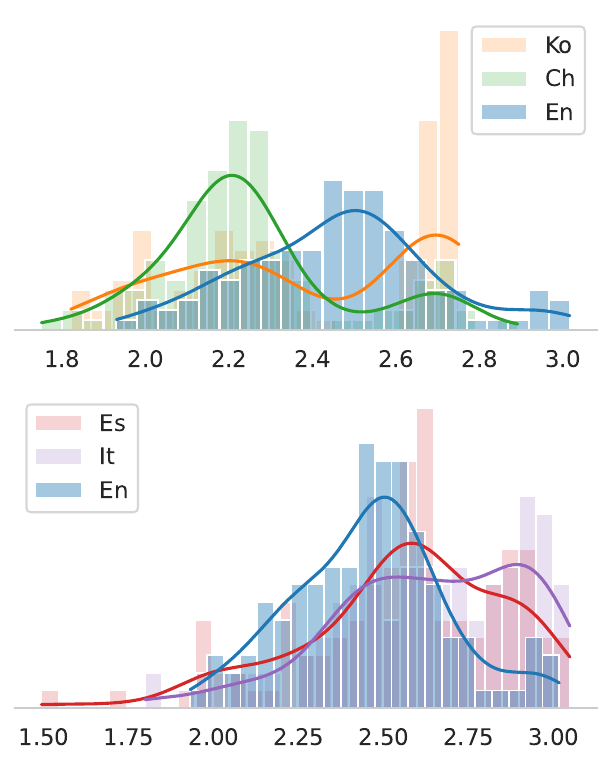}\label{subfig:olmo}}  
    \subfigure[SmolLM-1.7B]{\includegraphics[width=0.2455\textwidth]{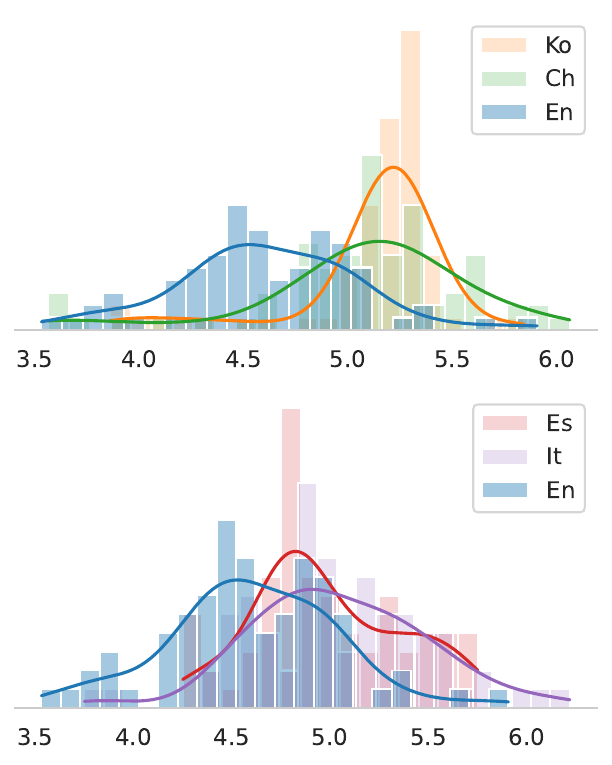}\label{subfig:smol}}
    \subfigure[Qwen2.5-3B]{\includegraphics[width=0.2455\textwidth]{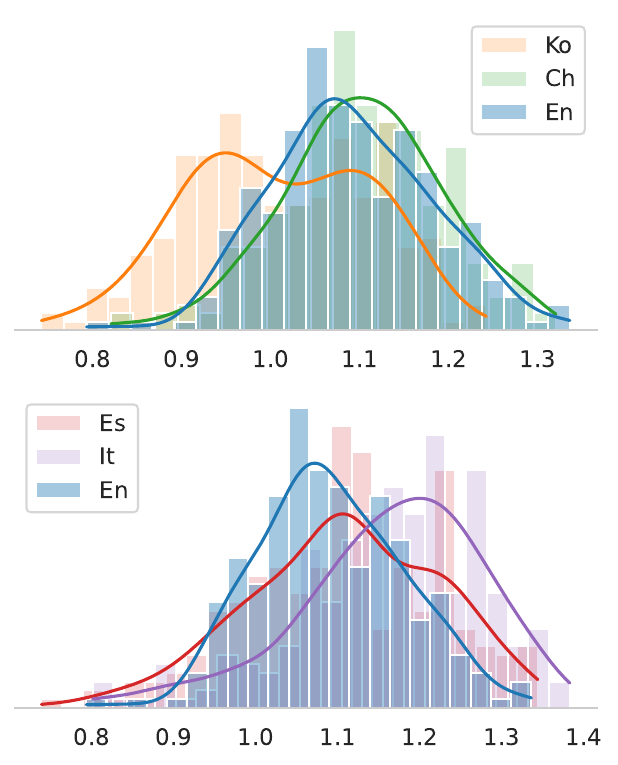}\label{subfig:qwen}}
    \subfigure[Llama-3.2-3B]{\includegraphics[width=0.2455\textwidth]{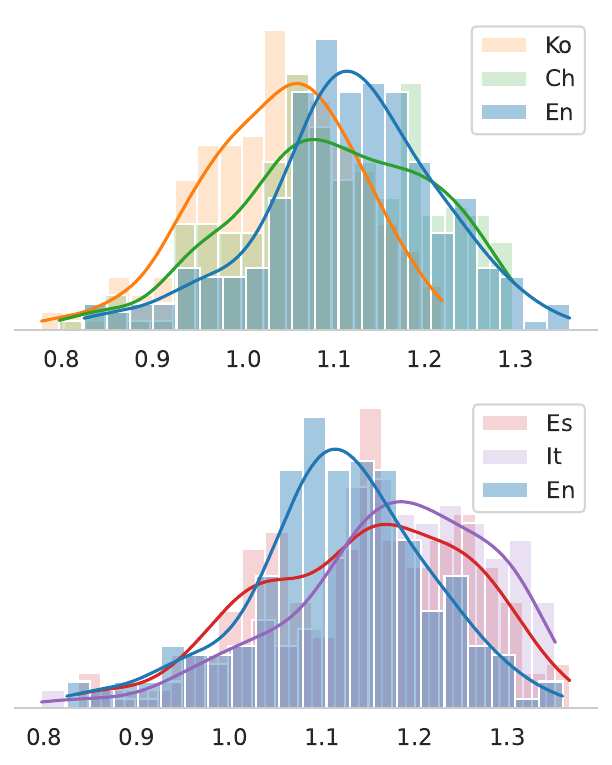}\label{subfig:llama}}  
    \vspace{-0.1in}
    \caption{Embedding norm distribution comparison between English and four other languages (2 non-Latin (top), 2 Latin (bottom)) across four language models: OLMo-1B and SmolLM-1.7B (monolingual pre-training) and Qwen2.5-3B and Llama-3.2-3B (multilingual pre-training). While English and non-English token embedding norm distributions of OLMo-1B and SmolLM-1.7B are distinct, they are similar in Qwen2.5-3B and Llama-3.2-3B.}
    \label{fig:norm}
\end{figure*}
\subsection{English preserves general representations}\label{subsec:represent}

\paragraph{Non-English reward modeling is detrimental to generalizability}In general, the generalizability of the downstream model is closely connected to \emph{how much the representations are preserved} during the fine-tuning \citep{aghajanyan2021better, razdaibiedina-etal-2023-representation}. We demonstrate this in RMs by ablating over different languages and tasks. We assess the general representation preservation of RMs used in Section \ref{sec:exp} by comparing their hidden states against the initial model. To do so, we measure how much the distinct representations are collapsed into similar spaces in Figure \ref{fig:emb}. In specific, we construct a matrix of the last hidden states $\mathcal{H}_\theta(x) \in \mathbb{R}^{5 \times d_\text{model}}$ across five languages using multilingual dataset BeleBele \citep{bandarkar-etal-2024-belebele}:
\begin{equation*}
    \mathcal{H}_\theta(x) = \text{concat}\left[ \left\{ H_\theta^l(x_l) \right\}_{l \in L} \right] \in \mathbb{R}^{|L| \times d_\text{model}},
\end{equation*}
where $H_\theta^l(x) \in \mathbb{R}^{d_\text{model}}$ refers to the last hidden state of the model $\theta$ for sequence $x_l$ in the language $l$, but with fixed context. Then, we measure the proportion of the largest singular value in $\mathcal{H}_\theta(x)$:
\begin{equation*}
    f_\theta(x) = \frac{\sigma_1}{\sum_{i=1}^{|L|}\sigma_i}, S = \text{diag}\left( \sigma_1, \ldots, \sigma_{|L|} \right),
\end{equation*}
with $S$ as singular value matrix of $\mathcal{H}_\theta(x)$. Intuitively, having $f_\theta(x)$ close to 1 imples the hidden states in different languages are homogeneous: \emph{i.e.,} representations are embedded into similar space.

In Figure \ref{fig:emb}, we plot $f_\theta(x)$ with different RMs. English RMs best preserve the representations by staying close to the base instruct model ("Inst"). On the other hand, Korean RMs ("Ko") tend to deviate the most from the base model, thereby homogenizing the multilingual representations the most. Both observations were more extreme in Llama-3.2-3B.

\paragraph{General representation preservation is crucial for cross-lingual/task transfer} Notably, the proclivity in general representation preservation in Figure \ref{fig:emb} aligns with the accuracy in Table \ref{tab:rb_full}. Non-English RMs with Llama-3.2-3B tend to introduce stronger representation collapse than Qwen2.5-3B in Figure \ref{fig:emb}. This aligns with Section \ref{subsec:results} as Llama-3.2-3B gets more severe degradation using target language RMs, implying the significance of representation preservation in cross-lingual transfer.

Furthermore, the same tendency holds for cross-task analysis. \texttt{RewardBench} has especially fine-grained divisions under the reasoning category (\emph{e.g.,} Java, Python, Rust, math) compared to other categories. Thus, strong generalization abilities are crucial to achieving decent scores in the reasoning category. Interestingly, English RMs dominate other languages in reasoning despite the fixed data across the languages in Table \ref{tab:rb_full}, which strongly supports the significance of representation preservation in cross-task generalization.

\subsection{MLM representations are language-aware}\label{subsec:emb}

In autoregressive language models \citep{radford2019language} with tied embeddings \citep{jiang2023mistral7b, gemmateam2024gemma2improvingopen}, the logits for next token is:
\begin{equation*}
    h_t \cdot E = \Bigl[ \| h_t \| \cdot \|e_i\| \cdot \cos \left(\theta_i\right)\Bigr]_{i=1}^{|V|},
\end{equation*}
where $\theta_i$ is the angle between $h_t$ and $e_i$. Therefore, the capability of language models in generative tasks is closely related to having \emph{good representations} \citep{edunov-etal-2019-pre} that could accurately align with the ideal next token. 

Token embeddings are a good proxy to understand the effectiveness of representations as they imply the imbalance in pre-training corpora \citep{chung2024stablelanguagemodelpretraining}, especially by \emph{linguality} in this study \citep{wen2023hyperpolyglot}. 
Thus, we can infer that language models with similar embedding norm distribution across the language will have decoder layers that can return language-aware fine-grained hidden states, which deserve to be preserved for their generalizability.

\paragraph{MLMs have similar token embedding norm distributions across the language}We validate this point by comparing the two models in Section \ref{sec:exp} with two monolingual pre-trained language models: OLMo-1B \citep{groeneveld-etal-2024-olmo} and SmolLM-1.7B \citep{allal2024SmolLM}. We clarify the lingualities in each model's pre-training in Appendix \ref{apdx:ling}.

We collect the disjoint language-specific token embedding norms for each model:
\begin{equation*}
    \mathbf{e}_l = \left\{ \|e_j\| \right\}_{j \in A_l}, A_l \subset V, \bigcap_{l \in L}A_l = \varnothing
\end{equation*}
where $A_l$ is the token indices of language $l$ in $V$. We compare $\mathbf{e}_{L}$ distribution over five languages. 

In Figure \ref{fig:norm}, the distribution for English in SmolLM-1.7B and OLMo-1B are distinct from four languages, especially Korean and Chinese, which are non-Latin languages that do not share similar alphabets. However, Qwen2.5-3B and Llama-3.2-3B have similar ranges and distributions across the languages, even in non-Latin languages.

Thus, we can infer that Qwen2.5-3B and Llama-3.2-3B, as MLMs, are sufficiently trained on the multilingual corpus to encode information with diverse linguality by having similar embedding norm distributions across the languages \citep{dagan2024getting, chung2024stablelanguagemodelpretraining}. This supports why representation preservation is a crucial condition for generalizable RMs with MLMs, as discussed in Section \ref{subsec:represent}.

\section{Multilingual Alignment using RM}\label{sec:align}
In this section, we perform experiments to outline the effects of using the reward models (RMs) from Section \ref{sec:exp} and how their cross-lingual transfer can propagate to the actual alignment process.

\subsection{Experimental Details}
We sample 10k prompts from the cleaned UltraFeedback dataset \citep{notus2023, cui2024ultrafeedbackboostinglanguagemodels} and translate prompts across target languages. Then, we sample four responses per prompt with Qwen2.5-7B-Instruct \citep{qwen25} and label them with desired RMs. By selecting the responses with the highest and lowest rewards, we prepare pairwise preference data. We train Qwen2.5-7B-Instruct on each language from the newly curated datasets with Direct Preference Optimization \citep[DPO]{rafailov2024direct}. Refer to Appendix \ref{apdx:config} for the detailed setup.

\paragraph{Evaluation} We evaluate the trained model's language-specific instruction-following capability with \texttt{Multilingual AlpacaEval}, adopted from the instances and evaluation pipeline of \texttt{AlpacaEval} \citep{alpaca_eval}. We report the detailed process and configurations in Appendix \ref{apdx:ae}.

\begin{figure}[t!]
    \centering
    \includegraphics[width=\columnwidth]{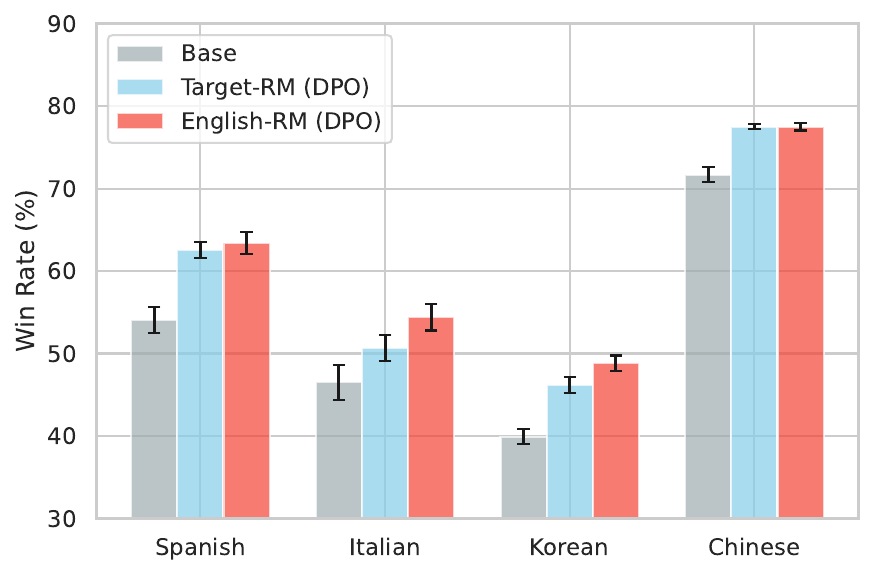}
    \vspace{-0.2in}
    \caption{\texttt{Multilingual AlpacaEval} results of Qwen2.5-7B-Instruct models fine-tuned with DPO on on-policy generations for four non-English languages over fine runs. The alignment data were labeled with either English RM or target language RM. Results are averaged over 5 runs.}
    \label{fig:main_ae}
\end{figure}

\subsection{Results and Analysis}
\paragraph{English RM largely improves base models in \emph{every} language} As shown in Figure \ref{fig:main_ae}, models aligned with English RM show a notable leap compared to Qwen2.5-7B-Instruct ("Base"), by increasing up to 9.3\% point in Spanish. As the win rate was measured against GPT-4-Turbo, a strong proprietary language model, such enhancements strongly support the validity of using English RMs for multilingual alignment in desired languages.

\paragraph{Exploiting English RMs is a desirable choice in multilingual alignment} We emphasize that using high-quality English preference data of better accessibility is a decent choice, considering the efficiency and efficacy in real-world cases. In Figure \ref{fig:main_ae}, models aligned with English RM outperformed or at least on par with ones with target language RMs, tied only in Chinese. Thus, adopting an English RM for multilingual alignment is a cost-efficient yet performant alternative, discarding the need for scaled translations for the reward model.

\section{Cross-lingual Transfer of External RMs}

Along with the controlled comparisons in Section \ref{sec:exp}, we analyze the cross-lingual transfer in off-the-shelf models on the original \texttt{RewardBench} through \texttt{Multilingual RewardBench}. To ensure diversity in reward modeling schemes, we selected two classifier reward models (RMs), ArmoRM-8B \citep{ArmoRM} and OffsetBias-8B \citep{park2024offsetbias}, alongside two generative RMs, GPT-4o\footnote{\url{https://platform.openai.com/docs/models/gpt-4o}} and Self-Taught-Llama-70B \citep{wang2024self}.

\paragraph{Classifier RMs} Two classifier RMs are both trained on top of Llama3-8B-Instruct \citep{dubey2024llama3herdmodels}, which are based on multilingual pre-trained language models (MLMs) as discussed in Appendix \ref{apdx:ling}. As in Table \ref{tab:rb_full}, these RMs also demonstrate strong cross-lingual transfer in four languages, mostly exceeding 70\% accuracy across the board in Table \ref{table:rb_avg}.

\paragraph{Generative reward models} Interestingly, we can observe strong cross-lingual transfer in the generative RMs in Table \ref{table:rb_avg}, as in the classifier RMs. As discussed in Section \ref{subsec:emb}, fine-grained representation learning is a crucial component for having strong downstream generative abilities. While the extent of multilingual pre-training in GPT-4o is not verifiable, GPT-4o has the least decrement in non-English settings. Meantime, Self-Taught-Llama-70B with extensive multilingual pre-training demonstrates the strongest cross-lingual transfer, achieving the best accuracies in all four non-English \texttt{Multilingual RewardBench}.

\begin{table}[t!]
\vskip 0.15in
\begin{center}
\begin{small}
\begin{sc}
\resizebox{\columnwidth}{!}{%
\begin{tabular}{l|c|c|c|c|c}
\toprule
\textbf{Model} & \textbf{En} & \textbf{Es} & \textbf{It} & \textbf{Ko} & \textbf{Ch}  \\
\toprule
ArmoRM-8B &\textbf{90.4} &80.1 &78.9 &71.5 &69.6 \\ 
OffsetBias-8B  &89.4 &78.9 &{79.5}  &74.5 &{73.1}   \\ 
\midrule

GPT-4o$^\dagger$ &86.7 & {80.4} & 78.6 &{75.2} &72.1  \\
ST-L-70B* &{90.0} &\textbf{83.1} &\textbf{81.5} &\textbf{75.6} &\textbf{74.1} \\

\bottomrule
\end{tabular}
}
\end{sc}
\end{small}
\end{center}
\caption{Averaged \textsc{Multilingual RewardBench} results in two classifier RMs (top) and two generative RMs (bottom). Off-the-shelf RMs based on MLMs show strong cross-lingual transfer as in Table \ref{tab:rb_full}.}
\label{table:rb_avg}
\end{table}

\section*{Conclusion}

We empirically demonstrate English as a \emph{lingua franca} in reward modeling, given recent multilingual pre-trained language models (MLMs). We explain this with two consecutive arguments. First, English reward models (RMs) best preserve the representations of initial MLMs, while other languages induce representation collapse. Second, MLM representations inherently have a rich understanding of languages and tasks, making them valuable to preserve in downstream tasks. By extending our analysis to the off-the-shelf reward models, we show that using MLMs for reward modeling is crucial for eliciting strong cross-lingual transfer. Through strong cross-lingual transfer in English RMs, we establish a concrete foundation for exploiting English RMs for multilingual alignment.

\section*{Limitations}

To extend to more languages and evaluation benchmarks, we have mainly utilized a 3B LLM to train the reward model (RM) with only 86k instances. However, as outlined in Appendix \ref{apdx:rb}, the 3B RMs are on par with a state-of-the-art RM, ArmoRM, which was trained with over 550k instances. Future works on the effects of data size and mixture will provide an enhanced understanding of our work.

Also, in Section \ref{sec:align}, we use the AlpacaEval evaluation setup, which utilizes LLM-generated reference responses and LLM-as-a-Judge to select a winning response. Therefore, while we show a vast increase in post-training alignment, the process relies on the multilinguality of OpenAI models and the evaluation biases of the LLM-based evaluations outlined in \citealp{zheng2023judging}.

\bibliography{custom}

\begin{thebibliography}{61}
\providecommand{\natexlab}[1]{#1}

\bibitem[{Aghajanyan et~al.(2021)Aghajanyan, Shrivastava, Gupta, Goyal, Zettlemoyer, and Gupta}]{aghajanyan2021better}
Armen Aghajanyan, Akshat Shrivastava, Anchit Gupta, Naman Goyal, Luke Zettlemoyer, and Sonal Gupta. 2021.
\newblock \href {https://openreview.net/forum?id=OQ08SN70M1V} {Better fine-tuning by reducing representational collapse}.
\newblock In \emph{International Conference on Learning Representations}.

\bibitem[{Allal et~al.(2024)Allal, Lozhkov, Bakouch, von Werra, and Wolf}]{allal2024SmolLM}
Loubna~Ben Allal, Anton Lozhkov, Elie Bakouch, Leandro von Werra, and Thomas Wolf. 2024.
\newblock Smollm - blazingly fast and remarkably powerful.

\bibitem[{Bandarkar et~al.(2024)Bandarkar, Liang, Muller, Artetxe, Shukla, Husa, Goyal, Krishnan, Zettlemoyer, and Khabsa}]{bandarkar-etal-2024-belebele}
Lucas Bandarkar, Davis Liang, Benjamin Muller, Mikel Artetxe, Satya~Narayan Shukla, Donald Husa, Naman Goyal, Abhinandan Krishnan, Luke Zettlemoyer, and Madian Khabsa. 2024.
\newblock \href {https://aclanthology.org/2024.acl-long.44} {The belebele benchmark: a parallel reading comprehension dataset in 122 language variants}.
\newblock In \emph{Proceedings of the 62nd Annual Meeting of the Association for Computational Linguistics (Volume 1: Long Papers)}, pages 749--775, Bangkok, Thailand and virtual meeting. Association for Computational Linguistics.

\bibitem[{Bartolome et~al.(2023)Bartolome, Martin, and Vila}]{notus2023}
Alvaro Bartolome, Gabriel Martin, and Daniel Vila. 2023.
\newblock Notus.
\newblock \url{https://github.com/argilla-io/notus}.

\bibitem[{Bradley and Terry(1952)}]{bradley1952rank}
Ralph~Allan Bradley and Milton~E Terry. 1952.
\newblock Rank analysis of incomplete block designs: I. the method of paired comparisons.
\newblock \emph{Biometrika}, 39(3/4):324--345.

\bibitem[{Chi et~al.(2022)Chi, Huang, Dong, Ma, Zheng, Singhal, Bajaj, Song, Mao, Huang, and Wei}]{chi-etal-2022-xlm}
Zewen Chi, Shaohan Huang, Li~Dong, Shuming Ma, Bo~Zheng, Saksham Singhal, Payal Bajaj, Xia Song, Xian-Ling Mao, Heyan Huang, and Furu Wei. 2022.
\newblock \href {https://doi.org/10.18653/v1/2022.acl-long.427} {{XLM}-{E}: Cross-lingual language model pre-training via {ELECTRA}}.
\newblock In \emph{Proceedings of the 60th Annual Meeting of the Association for Computational Linguistics (Volume 1: Long Papers)}, pages 6170--6182, Dublin, Ireland. Association for Computational Linguistics.

\bibitem[{Christiano et~al.(2017)Christiano, Leike, Brown, Martic, Legg, and Amodei}]{NIPS2017_d5e2c0ad}
Paul~F Christiano, Jan Leike, Tom Brown, Miljan Martic, Shane Legg, and Dario Amodei. 2017.
\newblock \href {https://proceedings.neurips.cc/paper_files/paper/2017/file/d5e2c0adad503c91f91df240d0cd4e49-Paper.pdf} {Deep reinforcement learning from human preferences}.
\newblock In \emph{Advances in Neural Information Processing Systems}, volume~30. Curran Associates, Inc.

\bibitem[{Chung et~al.(2024)Chung, Hong, An, Thorne, and Yun}]{chung2024stablelanguagemodelpretraining}
Woojin Chung, Jiwoo Hong, Na~Min An, James Thorne, and Se-Young Yun. 2024.
\newblock \href {https://arxiv.org/abs/2409.07787} {Stable language model pre-training by reducing embedding variability}.
\newblock \emph{Preprint}, arXiv:2409.07787.

\bibitem[{Conneau et~al.(2020)Conneau, Khandelwal, Goyal, Chaudhary, Wenzek, Guzm{\'a}n, Grave, Ott, Zettlemoyer, and Stoyanov}]{conneau-etal-2020-unsupervised}
Alexis Conneau, Kartikay Khandelwal, Naman Goyal, Vishrav Chaudhary, Guillaume Wenzek, Francisco Guzm{\'a}n, Edouard Grave, Myle Ott, Luke Zettlemoyer, and Veselin Stoyanov. 2020.
\newblock \href {https://doi.org/10.18653/v1/2020.acl-main.747} {Unsupervised cross-lingual representation learning at scale}.
\newblock In \emph{Proceedings of the 58th Annual Meeting of the Association for Computational Linguistics}, pages 8440--8451, Online. Association for Computational Linguistics.

\bibitem[{Cui et~al.(2024)Cui, Yuan, Ding, Yao, He, Zhu, Ni, Xie, Xie, Lin, Liu, and Sun}]{cui2024ultrafeedbackboostinglanguagemodels}
Ganqu Cui, Lifan Yuan, Ning Ding, Guanming Yao, Bingxiang He, Wei Zhu, Yuan Ni, Guotong Xie, Ruobing Xie, Yankai Lin, Zhiyuan Liu, and Maosong Sun. 2024.
\newblock \href {https://arxiv.org/abs/2310.01377} {Ultrafeedback: Boosting language models with scaled ai feedback}.
\newblock \emph{Preprint}, arXiv:2310.01377.

\bibitem[{Dagan et~al.(2024)Dagan, Synnaeve, and Roziere}]{dagan2024getting}
Gautier Dagan, Gabriel Synnaeve, and Baptiste Roziere. 2024.
\newblock \href {https://openreview.net/forum?id=ZFYBnLljtT} {Getting the most out of your tokenizer for pre-training and domain adaptation}.
\newblock In \emph{Forty-first International Conference on Machine Learning}.

\bibitem[{Dai et~al.(2024)Dai, Pan, Sun, Ji, Xu, Liu, Wang, and Yang}]{safe-rlhf}
Josef Dai, Xuehai Pan, Ruiyang Sun, Jiaming Ji, Xinbo Xu, Mickel Liu, Yizhou Wang, and Yaodong Yang. 2024.
\newblock \href {https://openreview.net/forum?id=TyFrPOKYXw} {Safe rlhf: Safe reinforcement learning from human feedback}.
\newblock In \emph{The Twelfth International Conference on Learning Representations}.

\bibitem[{Dang et~al.(2024)Dang, Ahmadian, Marchisio, Kreutzer, {\"U}st{\"u}n, and Hooker}]{dang2024rlhf}
John Dang, Arash Ahmadian, Kelly Marchisio, Julia Kreutzer, Ahmet {\"U}st{\"u}n, and Sara Hooker. 2024.
\newblock \href {https://arxiv.org/abs/2407.02552} {Rlhf can speak many languages: Unlocking multilingual preference optimization for llms}.
\newblock \emph{arXiv preprint arXiv:2407.02552}.

\bibitem[{Dettmers et~al.(2022)Dettmers, Lewis, Shleifer, and Zettlemoyer}]{dettmers2022optimizers}
Tim Dettmers, Mike Lewis, Sam Shleifer, and Luke Zettlemoyer. 2022.
\newblock 8-bit optimizers via block-wise quantization.
\newblock \emph{9th International Conference on Learning Representations, ICLR}.

\bibitem[{Dettmers et~al.(2023)Dettmers, Pagnoni, Holtzman, and Zettlemoyer}]{NEURIPS2023_1feb8787}
Tim Dettmers, Artidoro Pagnoni, Ari Holtzman, and Luke Zettlemoyer. 2023.
\newblock \href {https://proceedings.neurips.cc/paper_files/paper/2023/file/1feb87871436031bdc0f2beaa62a049b-Paper-Conference.pdf} {Qlora: Efficient finetuning of quantized llms}.
\newblock In \emph{Advances in Neural Information Processing Systems}, volume~36, pages 10088--10115. Curran Associates, Inc.

\bibitem[{Devlin et~al.(2019)Devlin, Chang, Lee, and Toutanova}]{devlin-etal-2019-bert}
Jacob Devlin, Ming-Wei Chang, Kenton Lee, and Kristina Toutanova. 2019.
\newblock \href {https://doi.org/10.18653/v1/N19-1423} {{BERT}: Pre-training of deep bidirectional transformers for language understanding}.
\newblock In \emph{Proceedings of the 2019 Conference of the North {A}merican Chapter of the Association for Computational Linguistics: Human Language Technologies, Volume 1 (Long and Short Papers)}, pages 4171--4186, Minneapolis, Minnesota. Association for Computational Linguistics.

\bibitem[{Dubey et~al.(2024)Dubey, Jauhri, Pandey, Kadian, Al-Dahle, Letman, Mathur, Schelten, Yang, Fan, Goyal, Hartshorn, Yang, Mitra, Sravankumar, Korenev, Hinsvark, Rao, Zhang, and et~al}]{dubey2024llama3herdmodels}
Abhimanyu Dubey, Abhinav Jauhri, Abhinav Pandey, Abhishek Kadian, Ahmad Al-Dahle, Aiesha Letman, Akhil Mathur, Alan Schelten, Amy Yang, Angela Fan, Anirudh Goyal, Anthony Hartshorn, Aobo Yang, Archi Mitra, Archie Sravankumar, Artem Korenev, Arthur Hinsvark, Arun Rao, Aston Zhang, and Aurelien~Rodriguez et~al. 2024.
\newblock \href {https://arxiv.org/abs/2407.21783} {The llama 3 herd of models}.
\newblock \emph{Preprint}, arXiv:2407.21783.

\bibitem[{Edunov et~al.(2019)Edunov, Baevski, and Auli}]{edunov-etal-2019-pre}
Sergey Edunov, Alexei Baevski, and Michael Auli. 2019.
\newblock \href {https://doi.org/10.18653/v1/N19-1409} {Pre-trained language model representations for language generation}.
\newblock In \emph{Proceedings of the 2019 Conference of the North {A}merican Chapter of the Association for Computational Linguistics: Human Language Technologies, Volume 1 (Long and Short Papers)}, pages 4052--4059, Minneapolis, Minnesota. Association for Computational Linguistics.

\bibitem[{Ethayarajh et~al.(2022)Ethayarajh, Choi, and Swayamdipta}]{pmlr-v162-ethayarajh22a}
Kawin Ethayarajh, Yejin Choi, and Swabha Swayamdipta. 2022.
\newblock \href {https://proceedings.mlr.press/v162/ethayarajh22a.html} {Understanding dataset difficulty with $\mathcal{V}$-usable information}.
\newblock In \emph{Proceedings of the 39th International Conference on Machine Learning}, volume 162 of \emph{Proceedings of Machine Learning Research}, pages 5988--6008. PMLR.

\bibitem[{Gao et~al.(2023)Gao, Schulman, and Hilton}]{gao2023scaling}
Leo Gao, John Schulman, and Jacob Hilton. 2023.
\newblock Scaling laws for reward model overoptimization.
\newblock In \emph{International Conference on Machine Learning}, pages 10835--10866. PMLR.

\bibitem[{Groeneveld et~al.(2024)Groeneveld, Beltagy, Walsh, Bhagia, Kinney, Tafjord, Jha, Ivison, Magnusson, Wang, Arora, Atkinson, Authur, Chandu, Cohan, Dumas, Elazar, Gu, Hessel, Khot, Merrill, Morrison, Muennighoff, Naik, Nam, Peters, Pyatkin, Ravichander, Schwenk, Shah, Smith, Strubell, Subramani, Wortsman, Dasigi, Lambert, Richardson, Zettlemoyer, Dodge, Lo, Soldaini, Smith, and Hajishirzi}]{groeneveld-etal-2024-olmo}
Dirk Groeneveld, Iz~Beltagy, Evan Walsh, Akshita Bhagia, Rodney Kinney, Oyvind Tafjord, Ananya Jha, Hamish Ivison, Ian Magnusson, Yizhong Wang, Shane Arora, David Atkinson, Russell Authur, Khyathi Chandu, Arman Cohan, Jennifer Dumas, Yanai Elazar, Yuling Gu, Jack Hessel, Tushar Khot, William Merrill, Jacob Morrison, Niklas Muennighoff, Aakanksha Naik, Crystal Nam, Matthew Peters, Valentina Pyatkin, Abhilasha Ravichander, Dustin Schwenk, Saurabh Shah, William Smith, Emma Strubell, Nishant Subramani, Mitchell Wortsman, Pradeep Dasigi, Nathan Lambert, Kyle Richardson, Luke Zettlemoyer, Jesse Dodge, Kyle Lo, Luca Soldaini, Noah Smith, and Hannaneh Hajishirzi. 2024.
\newblock \href {https://doi.org/10.18653/v1/2024.acl-long.841} {{OLM}o: Accelerating the science of language models}.
\newblock In \emph{Proceedings of the 62nd Annual Meeting of the Association for Computational Linguistics (Volume 1: Long Papers)}, pages 15789--15809, Bangkok, Thailand. Association for Computational Linguistics.

\bibitem[{Gugger et~al.(2022)Gugger, Debut, Wolf, Schmid, Mueller, Mangrulkar, Sun, and Bossan}]{accelerate}
Sylvain Gugger, Lysandre Debut, Thomas Wolf, Philipp Schmid, Zachary Mueller, Sourab Mangrulkar, Marc Sun, and Benjamin Bossan. 2022.
\newblock Accelerate: Training and inference at scale made simple, efficient and adaptable.
\newblock \url{https://github.com/huggingface/accelerate}.

\bibitem[{Han et~al.(2024)Han, Rao, Ettinger, Jiang, Lin, Lambert, Choi, and Dziri}]{wildguard2024}
Seungju Han, Kavel Rao, Allyson Ettinger, Liwei Jiang, Bill~Yuchen Lin, Nathan Lambert, Yejin Choi, and Nouha Dziri. 2024.
\newblock \href {https://arxiv.org/abs/2406.18495} {Wildguard: Open one-stop moderation tools for safety risks, jailbreaks, and refusals of llms}.
\newblock \emph{Preprint}, arXiv:2406.18495.

\bibitem[{Hong et~al.(2024)Hong, Lee, and Thorne}]{hong2024orpo}
Jiwoo Hong, Noah Lee, and James Thorne. 2024.
\newblock Orpo: Monolithic preference optimization without reference model.
\newblock \emph{EMNLP}.

\bibitem[{Hsu et~al.(2024)Hsu, Dai, Kothapalli, Song, Tang, and Zhu}]{liger2024}
Pin-Lun Hsu, Yun Dai, Vignesh Kothapalli, Qingquan Song, Shao Tang, and Siyu Zhu. 2024.
\newblock \href {https://github.com/linkedin/Liger-Kernel} {Liger-kernel: Efficient triton kernels for llm training}.

\bibitem[{Huang et~al.(2024)Huang, Noukhovitch, Hosseini, Rasul, Wang, and Tunstall}]{huang2024the}
Shengyi Huang, Michael Noukhovitch, Arian Hosseini, Kashif Rasul, Weixun Wang, and Lewis Tunstall. 2024.
\newblock \href {https://openreview.net/forum?id=kHO2ZTa8e3} {The n+ implementation details of {RLHF} with {PPO}: A case study on {TL};{DR} summarization}.
\newblock In \emph{First Conference on Language Modeling}.

\bibitem[{Ji et~al.(2023)Ji, Liu, Dai, Pan, Zhang, Bian, Chen, Sun, Wang, and Yang}]{ji2023beavertails}
Jiaming Ji, Mickel Liu, Juntao Dai, Xuehai Pan, Chi Zhang, Ce~Bian, Boyuan Chen, Ruiyang Sun, Yizhou Wang, and Yaodong Yang. 2023.
\newblock \href {https://openreview.net/forum?id=g0QovXbFw3} {Beavertails: Towards improved safety alignment of {LLM} via a human-preference dataset}.
\newblock In \emph{Thirty-seventh Conference on Neural Information Processing Systems Datasets and Benchmarks Track}.

\bibitem[{Jiang et~al.(2023)Jiang, Sablayrolles, Mensch, Bamford, Chaplot, de~las Casas, Bressand, Lengyel, Lample, Saulnier, Lavaud, Lachaux, Stock, Scao, Lavril, Wang, Lacroix, and Sayed}]{jiang2023mistral7b}
Albert~Q. Jiang, Alexandre Sablayrolles, Arthur Mensch, Chris Bamford, Devendra~Singh Chaplot, Diego de~las Casas, Florian Bressand, Gianna Lengyel, Guillaume Lample, Lucile Saulnier, Lélio~Renard Lavaud, Marie-Anne Lachaux, Pierre Stock, Teven~Le Scao, Thibaut Lavril, Thomas Wang, Timothée Lacroix, and William~El Sayed. 2023.
\newblock \href {https://arxiv.org/abs/2310.06825} {Mistral 7b}.
\newblock \emph{Preprint}, arXiv:2310.06825.

\bibitem[{Lambert et~al.(2024)Lambert, Pyatkin, Morrison, Miranda, Lin, Chandu, Dziri, Kumar, Zick, Choi, Smith, and Hajishirzi}]{rewardbench}
Nathan Lambert, Valentina Pyatkin, Jacob Morrison, LJ~Miranda, Bill~Yuchen Lin, Khyathi~Raghavi Chandu, Nouha Dziri, Sachin Kumar, Tom Zick, Yejin Choi, Noah~A. Smith, and Hannaneh Hajishirzi. 2024.
\newblock \href {https://doi.org/10.48550/arXiv.2403.13787} {Rewardbench: Evaluating reward models for language modeling}.
\newblock \emph{CoRR}, abs/2403.13787.

\bibitem[{Li et~al.(2023)Li, Zhang, Dubois, Taori, Gulrajani, Guestrin, Liang, and Hashimoto}]{alpaca_eval}
Xuechen Li, Tianyi Zhang, Yann Dubois, Rohan Taori, Ishaan Gulrajani, Carlos Guestrin, Percy Liang, and Tatsunori~B. Hashimoto. 2023.
\newblock Alpacaeval: An automatic evaluator of instruction-following models.
\newblock \url{https://github.com/tatsu-lab/alpaca_eval}.

\bibitem[{Loshchilov and Hutter(2019)}]{loshchilov2018decoupled}
Ilya Loshchilov and Frank Hutter. 2019.
\newblock \href {https://openreview.net/forum?id=Bkg6RiCqY7} {Decoupled weight decay regularization}.
\newblock In \emph{International Conference on Learning Representations}.

\bibitem[{Meng et~al.(2024)Meng, Xia, and Chen}]{meng2024simpo}
Yu~Meng, Mengzhou Xia, and Danqi Chen. 2024.
\newblock Simpo: Simple preference optimization with a reference-free reward.
\newblock \emph{arXiv preprint arXiv:2405.14734}.

\bibitem[{Park et~al.(2024)Park, Jwa, Ren, Kim, and Choi}]{park2024offsetbias}
Junsoo Park, Seungyeon Jwa, Meiying Ren, Daeyoung Kim, and Sanghyuk Choi. 2024.
\newblock \href {https://arxiv.org/abs/2407.06551} {Offsetbias: Leveraging debiased data for tuning evaluators}.
\newblock \emph{Preprint}, arXiv:2407.06551.

\bibitem[{Penedo et~al.(2024)Penedo, Kydlíček, allal, Lozhkov, Mitchell, Raffel, Werra, and Wolf}]{penedo2024finewebdatasetsdecantingweb}
Guilherme Penedo, Hynek Kydlíček, Loubna~Ben allal, Anton Lozhkov, Margaret Mitchell, Colin Raffel, Leandro~Von Werra, and Thomas Wolf. 2024.
\newblock \href {https://arxiv.org/abs/2406.17557} {The fineweb datasets: Decanting the web for the finest text data at scale}.
\newblock \emph{Preprint}, arXiv:2406.17557.

\bibitem[{Radford et~al.(2019)Radford, Wu, Child, Luan, Amodei, Sutskever et~al.}]{radford2019language}
Alec Radford, Jeffrey Wu, Rewon Child, David Luan, Dario Amodei, Ilya Sutskever, et~al. 2019.
\newblock Language models are unsupervised multitask learners.
\newblock \emph{OpenAI blog}, 1(8):9.

\bibitem[{Rafailov et~al.(2024)Rafailov, Sharma, Mitchell, Manning, Ermon, and Finn}]{rafailov2024direct}
Rafael Rafailov, Archit Sharma, Eric Mitchell, Christopher~D Manning, Stefano Ermon, and Chelsea Finn. 2024.
\newblock Direct preference optimization: Your language model is secretly a reward model.
\newblock \emph{Advances in Neural Information Processing Systems}, 36.

\bibitem[{Rajbhandari et~al.(2020)Rajbhandari, Rasley, Ruwase, and He}]{ds3}
Samyam Rajbhandari, Jeff Rasley, Olatunji Ruwase, and Yuxiong He. 2020.
\newblock Zero: memory optimizations toward training trillion parameter models.
\newblock In \emph{Proceedings of the International Conference for High Performance Computing, Networking, Storage and Analysis}, SC '20. IEEE Press.

\bibitem[{Razdaibiedina et~al.(2023)Razdaibiedina, Khetan, Karnin, Khashabi, and Madan}]{razdaibiedina-etal-2023-representation}
Anastasia Razdaibiedina, Ashish Khetan, Zohar Karnin, Daniel Khashabi, and Vivek Madan. 2023.
\newblock \href {https://doi.org/10.18653/v1/2023.findings-emnlp.975} {Representation projection invariance mitigates representation collapse}.
\newblock In \emph{Findings of the Association for Computational Linguistics: EMNLP 2023}, pages 14638--14664, Singapore. Association for Computational Linguistics.

\bibitem[{Soldaini et~al.(2024)Soldaini, Kinney, Bhagia, Schwenk, Atkinson, Authur, Bogin, Chandu, Dumas, Elazar, Hofmann, Jha, Kumar, Lucy, Lyu, Lambert, Magnusson, Morrison, Muennighoff, Naik, Nam, Peters, Ravichander, Richardson, Shen, Strubell, Subramani, Tafjord, Walsh, Zettlemoyer, Smith, Hajishirzi, Beltagy, Groeneveld, Dodge, and Lo}]{soldaini-etal-2024-dolma}
Luca Soldaini, Rodney Kinney, Akshita Bhagia, Dustin Schwenk, David Atkinson, Russell Authur, Ben Bogin, Khyathi Chandu, Jennifer Dumas, Yanai Elazar, Valentin Hofmann, Ananya Jha, Sachin Kumar, Li~Lucy, Xinxi Lyu, Nathan Lambert, Ian Magnusson, Jacob Morrison, Niklas Muennighoff, Aakanksha Naik, Crystal Nam, Matthew Peters, Abhilasha Ravichander, Kyle Richardson, Zejiang Shen, Emma Strubell, Nishant Subramani, Oyvind Tafjord, Evan Walsh, Luke Zettlemoyer, Noah Smith, Hannaneh Hajishirzi, Iz~Beltagy, Dirk Groeneveld, Jesse Dodge, and Kyle Lo. 2024.
\newblock \href {https://doi.org/10.18653/v1/2024.acl-long.840} {Dolma: an open corpus of three trillion tokens for language model pretraining research}.
\newblock In \emph{Proceedings of the 62nd Annual Meeting of the Association for Computational Linguistics (Volume 1: Long Papers)}, pages 15725--15788, Bangkok, Thailand. Association for Computational Linguistics.

\bibitem[{Son et~al.(2024)Son, Ko, Lee, Kim, and Hong}]{son2024llm}
Guijin Son, Hyunwoo Ko, Hoyoung Lee, Yewon Kim, and Seunghyeok Hong. 2024.
\newblock Llm-as-a-judge \& reward model: What they can and cannot do.
\newblock \emph{arXiv preprint arXiv:2409.11239}.

\bibitem[{Stiennon et~al.(2020)Stiennon, Ouyang, Wu, Ziegler, Lowe, Voss, Radford, Amodei, and Christiano}]{NEURIPS2020_1f89885d}
Nisan Stiennon, Long Ouyang, Jeffrey Wu, Daniel Ziegler, Ryan Lowe, Chelsea Voss, Alec Radford, Dario Amodei, and Paul~F Christiano. 2020.
\newblock \href {https://proceedings.neurips.cc/paper_files/paper/2020/file/1f89885d556929e98d3ef9b86448f951-Paper.pdf} {Learning to summarize with human feedback}.
\newblock In \emph{Advances in Neural Information Processing Systems}, volume~33, pages 3008--3021. Curran Associates, Inc.

\bibitem[{Team(2024{\natexlab{a}})}]{gemmateam2024gemma2improvingopen}
Gemma Team. 2024{\natexlab{a}}.
\newblock \href {https://arxiv.org/abs/2408.00118} {Gemma 2: Improving open language models at a practical size}.
\newblock \emph{Preprint}, arXiv:2408.00118.

\bibitem[{Team(2024{\natexlab{b}})}]{qwen25}
Qwen Team. 2024{\natexlab{b}}.
\newblock \href {https://qwenlm.github.io/blog/qwen2.5/} {Qwen2.5: A party of foundation models}.

\bibitem[{{\"U}st{\"u}n et~al.(2024){\"U}st{\"u}n, Aryabumi, Yong, Ko, D{'}souza, Onilude, Bhandari, Singh, Ooi, Kayid, Vargus, Blunsom, Longpre, Muennighoff, Fadaee, Kreutzer, and Hooker}]{ustun-etal-2024-aya}
Ahmet {\"U}st{\"u}n, Viraat Aryabumi, Zheng Yong, Wei-Yin Ko, Daniel D{'}souza, Gbemileke Onilude, Neel Bhandari, Shivalika Singh, Hui-Lee Ooi, Amr Kayid, Freddie Vargus, Phil Blunsom, Shayne Longpre, Niklas Muennighoff, Marzieh Fadaee, Julia Kreutzer, and Sara Hooker. 2024.
\newblock \href {https://doi.org/10.18653/v1/2024.acl-long.845} {Aya model: An instruction finetuned open-access multilingual language model}.
\newblock In \emph{Proceedings of the 62nd Annual Meeting of the Association for Computational Linguistics (Volume 1: Long Papers)}, pages 15894--15939, Bangkok, Thailand. Association for Computational Linguistics.

\bibitem[{von Werra et~al.(2020)von Werra, Belkada, Tunstall, Beeching, Thrush, Lambert, Huang, Rasul, and Gallouédec}]{vonwerra2022trl}
Leandro von Werra, Younes Belkada, Lewis Tunstall, Edward Beeching, Tristan Thrush, Nathan Lambert, Shengyi Huang, Kashif Rasul, and Quentin Gallouédec. 2020.
\newblock Trl: Transformer reinforcement learning.
\newblock \url{https://github.com/huggingface/trl}.

\bibitem[{Wang et~al.(2024{\natexlab{a}})Wang, Zheng, Chen, Liu, Dou, Huang, Shen, Jin, Zhou, Shi, Gao, Xu, Zhou, Fan, Xi, Zhao, Wang, Ji, Yan, Shen, Chen, Gui, Zhang, Qiu, Huang, Wu, and Jiang}]{secrets2rm}
Binghai Wang, Rui Zheng, Lu~Chen, Yan Liu, Shihan Dou, Caishuang Huang, Wei Shen, Senjie Jin, Enyu Zhou, Chenyu Shi, Songyang Gao, Nuo Xu, Yuhao Zhou, Xiaoran Fan, Zhiheng Xi, Jun Zhao, Xiao Wang, Tao Ji, Hang Yan, Lixing Shen, Zhan Chen, Tao Gui, Qi~Zhang, Xipeng Qiu, Xuanjing Huang, Zuxuan Wu, and Yu-Gang Jiang. 2024{\natexlab{a}}.
\newblock \href {https://doi.org/10.48550/arXiv.2401.06080} {Secrets of rlhf in large language models part ii: Reward modeling}.
\newblock \emph{CoRR}, abs/2401.06080.

\bibitem[{Wang et~al.(2024{\natexlab{b}})Wang, Xiong, Xie, Zhao, and Zhang}]{ArmoRM}
Haoxiang Wang, Wei Xiong, Tengyang Xie, Han Zhao, and Tong Zhang. 2024{\natexlab{b}}.
\newblock Interpretable preferences via multi-objective reward modeling and mixture-of-experts.
\newblock \emph{arXiv preprint arXiv:2406.12845}.

\bibitem[{Wang et~al.(2024{\natexlab{c}})Wang, Minervini, and Ponti}]{wang-etal-2024-probing-emergence}
Hetong Wang, Pasquale Minervini, and Edoardo Ponti. 2024{\natexlab{c}}.
\newblock \href {https://doi.org/10.18653/v1/2024.findings-acl.724} {Probing the emergence of cross-lingual alignment during {LLM} training}.
\newblock In \emph{Findings of the Association for Computational Linguistics ACL 2024}, pages 12159--12173, Bangkok, Thailand and virtual meeting. Association for Computational Linguistics.

\bibitem[{Wang et~al.(2024{\natexlab{d}})Wang, Kulikov, Golovneva, Yu, Yuan, Dwivedi-Yu, Pang, Fazel-Zarandi, Weston, and Li}]{wang2024self}
Tianlu Wang, Ilia Kulikov, Olga Golovneva, Ping Yu, Weizhe Yuan, Jane Dwivedi-Yu, Richard~Yuanzhe Pang, Maryam Fazel-Zarandi, Jason Weston, and Xian Li. 2024{\natexlab{d}}.
\newblock Self-taught evaluators.
\newblock \emph{arXiv preprint arXiv:2408.02666}.

\bibitem[{Wang et~al.(2024{\natexlab{e}})Wang, Dong, Delalleau, Zeng, Shen, Egert, Zhang, Sreedhar, and Kuchaiev}]{wang2024helpsteer2opensourcedatasettraining}
Zhilin Wang, Yi~Dong, Olivier Delalleau, Jiaqi Zeng, Gerald Shen, Daniel Egert, Jimmy~J. Zhang, Makesh~Narsimhan Sreedhar, and Oleksii Kuchaiev. 2024{\natexlab{e}}.
\newblock \href {https://arxiv.org/abs/2406.08673} {Helpsteer2: Open-source dataset for training top-performing reward models}.
\newblock \emph{Preprint}, arXiv:2406.08673.

\bibitem[{Wang et~al.(2024{\natexlab{f}})Wang, Dong, Zeng, Adams, Sreedhar, Egert, Delalleau, Scowcroft, Kant, Swope, and Kuchaiev}]{wang-etal-2024-helpsteer}
Zhilin Wang, Yi~Dong, Jiaqi Zeng, Virginia Adams, Makesh~Narsimhan Sreedhar, Daniel Egert, Olivier Delalleau, Jane Scowcroft, Neel Kant, Aidan Swope, and Oleksii Kuchaiev. 2024{\natexlab{f}}.
\newblock \href {https://doi.org/10.18653/v1/2024.naacl-long.185} {{H}elp{S}teer: Multi-attribute helpfulness dataset for {S}teer{LM}}.
\newblock In \emph{Proceedings of the 2024 Conference of the North American Chapter of the Association for Computational Linguistics: Human Language Technologies (Volume 1: Long Papers)}, pages 3371--3384, Mexico City, Mexico. Association for Computational Linguistics.

\bibitem[{Wen-Yi and Mimno(2023)}]{wen2023hyperpolyglot}
Andrea~W Wen-Yi and David Mimno. 2023.
\newblock \href {https://doi.org/10.18653/v1/2023.emnlp-main.71} {Hyperpolyglot {LLM}s: Cross-lingual interpretability in token embeddings}.
\newblock In \emph{Proceedings of the 2023 Conference on Empirical Methods in Natural Language Processing}, pages 1124--1131, Singapore. Association for Computational Linguistics.

\bibitem[{Wu et~al.(2024)Wu, Balashankar, Kim, Eisenstein, and Beirami}]{wu2024reuse}
Zhaofeng Wu, Ananth Balashankar, Yoon Kim, Jacob Eisenstein, and Ahmad Beirami. 2024.
\newblock \href {https://openreview.net/forum?id=IUhNTCNjox} {Reuse your rewards: Reward model transfer for zero-shot cross-lingual alignment}.
\newblock In \emph{ICML 2024 Workshop on Models of Human Feedback for AI Alignment}.

\bibitem[{Xu et~al.(2024{\natexlab{a}})Xu, Murray, Koehn, Hoang, Eriguchi, and Khayrallah}]{xu2024xalmaplugplay}
Haoran Xu, Kenton Murray, Philipp Koehn, Hieu Hoang, Akiko Eriguchi, and Huda Khayrallah. 2024{\natexlab{a}}.
\newblock \href {https://arxiv.org/abs/2410.03115} {X-alma: Plug \& play modules and adaptive rejection for quality translation at scale}.
\newblock \emph{Preprint}, arXiv:2410.03115.

\bibitem[{Xu et~al.(2024{\natexlab{b}})Xu, Jiang, Niu, Deng, Poovendran, Choi, and Lin}]{xu2024magpiealignmentdatasynthesis}
Zhangchen Xu, Fengqing Jiang, Luyao Niu, Yuntian Deng, Radha Poovendran, Yejin Choi, and Bill~Yuchen Lin. 2024{\natexlab{b}}.
\newblock \href {https://arxiv.org/abs/2406.08464} {Magpie: Alignment data synthesis from scratch by prompting aligned llms with nothing}.
\newblock \emph{Preprint}, arXiv:2406.08464.

\bibitem[{Xue et~al.(2021)Xue, Constant, Roberts, Kale, Al-Rfou, Siddhant, Barua, and Raffel}]{xue-etal-2021-mt5}
Linting Xue, Noah Constant, Adam Roberts, Mihir Kale, Rami Al-Rfou, Aditya Siddhant, Aditya Barua, and Colin Raffel. 2021.
\newblock \href {https://doi.org/10.18653/v1/2021.naacl-main.41} {m{T}5: A massively multilingual pre-trained text-to-text transformer}.
\newblock In \emph{Proceedings of the 2021 Conference of the North American Chapter of the Association for Computational Linguistics: Human Language Technologies}, pages 483--498, Online. Association for Computational Linguistics.

\bibitem[{Yang et~al.(2024)Yang, Yang, Hui, Zheng, Yu, Zhou, Li, Li, Liu, Huang, Dong, Wei, Lin, Tang, Wang, Yang, Tu, Zhang, Ma, Xu, Zhou, Bai, He, Lin, Dang, Lu, Chen, Yang, Li, Xue, Ni, Zhang, Wang, Peng, Men, Gao, Lin, Wang, Bai, Tan, Zhu, Li, Liu, Ge, Deng, Zhou, Ren, Zhang, Wei, Ren, Fan, Yao, Zhang, Wan, Chu, Liu, Cui, Zhang, and Fan}]{qwen2}
An~Yang, Baosong Yang, Binyuan Hui, Bo~Zheng, Bowen Yu, Chang Zhou, Chengpeng Li, Chengyuan Li, Dayiheng Liu, Fei Huang, Guanting Dong, Haoran Wei, Huan Lin, Jialong Tang, Jialin Wang, Jian Yang, Jianhong Tu, Jianwei Zhang, Jianxin Ma, Jin Xu, Jingren Zhou, Jinze Bai, Jinzheng He, Junyang Lin, Kai Dang, Keming Lu, Keqin Chen, Kexin Yang, Mei Li, Mingfeng Xue, Na~Ni, Pei Zhang, Peng Wang, Ru~Peng, Rui Men, Ruize Gao, Runji Lin, Shijie Wang, Shuai Bai, Sinan Tan, Tianhang Zhu, Tianhao Li, Tianyu Liu, Wenbin Ge, Xiaodong Deng, Xiaohuan Zhou, Xingzhang Ren, Xinyu Zhang, Xipin Wei, Xuancheng Ren, Yang Fan, Yang Yao, Yichang Zhang, Yu~Wan, Yunfei Chu, Yuqiong Liu, Zeyu Cui, Zhenru Zhang, and Zhihao Fan. 2024.
\newblock Qwen2 technical report.
\newblock \emph{arXiv preprint arXiv:2407.10671}.

\bibitem[{Zhang et~al.(2024)Zhang, Lee, Fang, Yu, Jia, Jiang, and Barbieri}]{zhang-etal-2024-plug}
Zhihan Zhang, Dong-Ho Lee, Yuwei Fang, Wenhao Yu, Mengzhao Jia, Meng Jiang, and Francesco Barbieri. 2024.
\newblock \href {https://doi.org/10.18653/v1/2024.acl-long.379} {{PLUG}: Leveraging pivot language in cross-lingual instruction tuning}.
\newblock In \emph{Proceedings of the 62nd Annual Meeting of the Association for Computational Linguistics (Volume 1: Long Papers)}, pages 7025--7046, Bangkok, Thailand. Association for Computational Linguistics.

\bibitem[{Zheng et~al.(2023)Zheng, Chiang, Sheng, Zhuang, Wu, Zhuang, Lin, Li, Li, Xing et~al.}]{zheng2023judging}
Lianmin Zheng, Wei-Lin Chiang, Ying Sheng, Siyuan Zhuang, Zhanghao Wu, Yonghao Zhuang, Zi~Lin, Zhuohan Li, Dacheng Li, Eric Xing, et~al. 2023.
\newblock Judging llm-as-a-judge with mt-bench and chatbot arena.
\newblock \emph{Advances in Neural Information Processing Systems}, 36:46595--46623.

\bibitem[{Zhu et~al.(2024)Zhu, Frick, Wu, Zhu, Ganesan, Chiang, Zhang, and Jiao}]{zhu2024starlingb}
Banghua Zhu, Evan Frick, Tianhao Wu, Hanlin Zhu, Karthik Ganesan, Wei-Lin Chiang, Jian Zhang, and Jiantao Jiao. 2024.
\newblock \href {https://openreview.net/forum?id=GqDntYTTbk} {Starling-7b: Improving helpfulness and harmlessness with {RLAIF}}.
\newblock In \emph{First Conference on Language Modeling}.

\bibitem[{Ziegler et~al.(2020)Ziegler, Stiennon, Wu, Brown, Radford, Amodei, Christiano, and Irving}]{ziegler2020finetuninglanguagemodelshuman}
Daniel~M. Ziegler, Nisan Stiennon, Jeffrey Wu, Tom~B. Brown, Alec Radford, Dario Amodei, Paul Christiano, and Geoffrey Irving. 2020.
\newblock \href {https://arxiv.org/abs/1909.08593} {Fine-tuning language models from human preferences}.
\newblock \emph{Preprint}, arXiv:1909.08593.

\end{thebibliography}

\appendix

\onecolumn

\section{Data Curation}\label{apdx:data}

We used full datasets for HelpSteer2, SafeRLHF, and Offsetbias. We filtered the prompts with one harmful and unharmful response each for WildGuard, finally having 8,383 instances. Lastly, we randomly sample 60,000 instances from the synthetic preference dataset comprising responses from Llama-3-70B-Instruct \citep{dubey2024llama3herdmodels} and Gemma-2-9B-It \citep{gemmateam2024gemma2improvingopen} labeled with ArmoRM \citep{ArmoRM}. From the 108k instances, we finally select 80\% of instances as the train set.

\section{Training Configurations} \label{apdx:config}

Both reward modeling and downstream on-policy preference optimization were down using Hugging Face TRL library \citep{vonwerra2022trl} on 4 NVIDIA A100 GPUs with Accelerate \citep{accelerate} and DeepSpeed ZeRO 3 \citep{ds3}, and Paged AdamW optimizer \citep{loshchilov2018decoupled, NEURIPS2023_1feb8787} with 8-bit precision \citep{dettmers2022optimizers}.

\subsection{Reward Modeling}

We used a maximum learning rate of $1e-5$ and 10\% of warm-up followed by cosine decay. The projection head for the reward model was initialized with $\mathcal{N}\left(0, 1/\sqrt{d_\text{model} + 1} \right)$ \citep{NEURIPS2020_1f89885d, huang2024the}. The global batch was set to 128.

\subsection{On-Policy Preference Optimization}

We fine-tune Qwen2.5-7B-Instruct \citep{qwen25} with DPO using Liger-kernel \citep{liger2024}. We use a cosine decaying learning rate scheduler for single epoch training.
\paragraph{DPO configurations}We apply $\beta=0.1$ with the learning rate of $5\text{e}-7$. The global batch size was set to 32 using gradient accumulation steps of 8 with a per-device batch size of 1, which was the maximum number for NVIDIA A100 80GiB.
\paragraph{Data curation}To construct the preference pairs for preference optimization, we sample 4 responses from Qwen-2.5-7B-Instruct. Then, we compute the rewards through the reward models and select the response with the highest and lowest reward values as the preference pairs for training the checkpoints through DPO.

\section{Linguality in Pre-training}\label{apdx:ling}
Olmo-1B and SmolLM-1.7B are selectively pre-trained on Dolma \citep{soldaini-etal-2024-dolma} and an English-focused subset of FineWeb \citep{penedo2024finewebdatasetsdecantingweb}, respectively: \emph{i.e.,} monolingual pre-training. On the other hand, the Qwen2.5 series is pre-trained on more than 7 trillion tokens comprising more than 30 languages \citep{qwen2, qwen25}: \emph{i.e.,} multilingual pre-training. Similarly, 8\% of 15 trillion tokens for pre-training Llama-3 series were multilingual \citep{dubey2024llama3herdmodels}.

\section{\textsc{Multilingual AlpacaEval} Setup} \label{apdx:ae}

Starting from the 805 translated prompt instances\footnote{\url{https://huggingface.co/datasets/zhihz0535/X-AlpacaEval}} \citep{zhang-etal-2024-plug}, we compute the language-specific win-rate of the model evaluated by GPT-4o\footnote{\url{https://platform.openai.com/docs/models/gpt-4o}} against the reference responses from GPT-4-Turbo\footnote{\href{https://platform.openai.com/docs/models/gpt-4-turbo-and-gpt-4}{https://platform.openai.com/docs/models/gpt-4-turbo-and-gpt-4}}. Given the generations from the reference model and aligned model, we adopt a LLM-as-a-Judge evaluation given the evaluation template\footnote{\href{https://github.com/tatsu-lab/alpaca_eval/blob/main/src/alpaca_eval/evaluators_configs/alpaca_eval_clf_gpt4_turbo/alpaca_eval_clf.txt}{https://github.com/tatsu-lab/alpaca\_eval/blob/main/src/alpaca\_eval/evaluators\_configs/alpaca\_eval\_clf\_gpt4\_turbo/alpaca\_eval\_clf.txt}}.

\clearpage

\section{\textsc{RewardBench} Evaluation Results Across Languages} \label{apdx:rb}

\begin{table*}[ht!]
\vskip 0.15in
\begin{center}
\begin{small}
\begin{sc}
\begin{tabular}{l|c|c|c|c|c}
\toprule
\textbf{Reward Model} & \textbf{Chat} & \textbf{Chat(H)} & \textbf{Safety} & \textbf{Reason} & \textbf{Avg.} \\
\toprule
ArmoRM-L3-8B* & 96.9 & 76.8 & 90.5 & 97.3 & 90.4 \\ 
\midrule

L32-3B-IT-En & 92.5  &  81.8  &  90.2  &  95.5  &  90.0   \\
L32-3B-IT-Sp & 82.1  &  71.7  &  88.2  &  81.5  &  80.9\\
L32-3B-IT-It & 86.3  &  66.0  &  88.4  &  75.4  &  79.0\\
L32-3B-IT-Ko & 84.4  &  70.6  &  84.8  &  78.7  &  79.6 \\
L32-3B-IT-Ch & 82.4  &  69.7  &  85.5  &  86.6  &  81.0 \\
\midrule
Q25-3B-IT-En & 89.1  &  75.2  &  87.3  &  95.4  &  86.8 \\
Q25-3B-IT-Sp & 89.7  &  70.4  &  85.1  &  83.2  &  82.1  \\
Q25-3B-IT-It & 88.3  &  68.9  &  86.2  &  88.8  &  83.0 \\
Q25-3B-IT-Ko & 86.3  &  69.5  &  84.6  &  76.8  &  79.3  \\
Q25-3B-IT-Ch & 84.6  &  68.2  &  84.8  &  89.1  &  81.7 \\
\midrule
Q25-7B-IT-En & 91.3  &  81.6  &  90.3  &  96.5  &  89.9  \\
Q25-7B-IT-Sp & 90.5  &  75.9  &  89.5  &  94.1  &  87.5  \\
Q25-7B-IT-It & 90.8  &  74.1  &  88.5  &  92.5  &  86.5   \\
Q25-7B-IT-Ko & 89.4  &  70.8  &  87.9  &  94.9  &  85.8   \\
Q25-7B-IT-Ch & 83.2  &  72.6  &  87.2  &  90.8  &  83.5  \\
\bottomrule
\end{tabular}
\end{sc}
\end{small}
\end{center}
\caption{\textsc{RewardBench} results for reward model comparison across four different categories. (* denotes off-the-shelf models)}
\label{table:apdx_rb}
\vskip -0.1in
\end{table*}


\begin{table*}[ht!]
\vskip 0.15in
\begin{center}
\begin{small}
\begin{sc}
\begin{tabular}{l|c|c|c|c|c}
\toprule
\textbf{Reward Model} & \textbf{Chat} & \textbf{Chat(H)} & \textbf{Safety} & \textbf{Reason} & \textbf{Avg.} \\
\toprule
ArmoRM-L3-8B* & 89.4  &  64.5  &  89.0  &  77.5  &  80.1 \\ 
\midrule
L32-3B-IT-En & 86.3  &  69.3  &  89.3  &  72.4  &  79.3\\
L32-3B-IT-Sp & 79.1  &  67.3  &  88.0  &  65.5  &  75.0\\
L32-3B-IT-It & 80.4  &  63.2  &  88.0  &  64.8  &  74.1 \\
L32-3B-IT-Ko & 79.1  &  63.8  &  84.0  &  54.8  &  70.4 \\
L32-3B-IT-Ch & 77.9  &  64.9  &  84.1  &  59.4  &  71.6 \\
\midrule
Q25-3B-IT-En & 82.7  &  68.0  &  88.3  &  73.6  &  78.1\\
Q25-3B-IT-Sp & 80.7  &  68.2  &  84.8  &  68.2  &  75.5 \\
Q25-3B-IT-It & 78.2  &  67.5  &  87.0  &  73.4  &  76.6  \\
Q25-3B-IT-Ko & 77.1  &  67.1  &  85.3  &  58.4  &  72.0  \\
Q25-3B-IT-Ch & 78.8  &  64.5  &  85.3  &  76.4  &  76.2  \\
\midrule
Q25-7B-IT-En & 82.1  &  73.7  &  91.4  &  73.3  &  80.1\\
Q25-7B-IT-Sp & 84.1  &  71.5  &  89.9  &  78.4  &  81.0   \\
Q25-7B-IT-It & 84.6  &  70.0  &  89.2  &  78.3  &  80.5  \\
Q25-7B-IT-Ko & 84.9  &  65.8  &  87.0  &  76.0  &  78.4   \\
Q25-7B-IT-Ch & 83.5  &  66.0  &  87.2  &  69.5  &  76.5   \\
\bottomrule
\end{tabular}
\end{sc}
\end{small}
\end{center}
\caption{Spanish \textsc{RewardBench} results for reward model comparison across four different categories. (* denotes off-the-shelf models)}
\label{table:rb_sp}
\vskip -0.1in
\end{table*}

\begin{table*}[ht!]
\vskip 0.15in
\begin{center}
\begin{small}
\begin{sc}
\begin{tabular}{l|c|c|c|c|c}
\toprule
\textbf{Reward Model} & \textbf{Chat} & \textbf{Chat(H)} & \textbf{Safety} & \textbf{Reason} & \textbf{Avg.} \\
\toprule
ArmoRM-L3-8B* & 83.2  &  65.4  &  88.6  &  78.5  &  78.9  \\ 
\midrule
L32-3B-IT-En & 83.0  &  69.3  &  88.7  &  75.1  &  79.0 \\
L32-3B-IT-Sp & 74.9  &  67.8  &  87.6  &  65.7  &  74.0 \\
L32-3B-IT-It & 75.4  &  62.5  &  88.5  &  65.7  &  73.0 \\
L32-3B-IT-Ko & 77.7  &  64.9  &  84.8  &  57.1  &  71.1 \\
L32-3B-IT-Ch & 75.4  &  62.5  &  84.5  &  61.7  &  71.0 \\
\midrule
Q25-3B-IT-En & 83.2  &  68.2  &  88.4  &  76.0  &  79.0\\
Q25-3B-IT-Sp & 81.0  &  65.8  &  84.3  &  70.9  &  75.5\\
Q25-3B-IT-It & 77.1  &  67.8  &  85.7  &  72.8  &  75.8 \\
Q25-3B-IT-Ko & 78.8  &  68.0  &  82.5  &  61.7  &  72.7 \\
Q25-3B-IT-Ch & 82.1  &  64.9  &  83.7  &  76.7  &  76.9\\
\midrule
Q25-7B-IT-En & 82.4  &  73.0  &  89.6  &  75.1  &  80.0  \\
Q25-7B-IT-Sp & 84.6  &  69.3  &  89.1  &  79.8  &  80.7   \\
Q25-7B-IT-It & 80.2  &  69.7  &  87.9  &  78.5  &  79.1   \\
Q25-7B-IT-Ko & 84.1  &  64.3  &  85.8  &  72.7  &  76.7   \\
Q25-7B-IT-Ch & 81.8  &  65.8  &  86.5  &  67.9  &  75.5   \\
\bottomrule
\end{tabular}
\end{sc}
\end{small}
\end{center}
\caption{Italian \textsc{RewardBench} results for reward model comparison across four different categories. (* denotes off-the-shelf models)}
\label{table:rb_it}
\vskip -0.1in
\end{table*}

\begin{table*}[ht!]
\vskip 0.15in
\begin{center}
\begin{small}
\begin{sc}
\begin{tabular}{l|c|c|c|c|c}
\toprule
\textbf{Reward Model} & \textbf{Chat} & \textbf{Chat(H)} & \textbf{Safety} & \textbf{Reason} & \textbf{Avg.} \\
\toprule
ArmoRM-L3-8B* & 66.5  &  60.3  &  83.8  &  75.3  &  71.5 \\ 
\midrule
L32-3B-IT-En & 69.8  &  59.4  &  84.3  &  73.0  &  71.6\\
L32-3B-IT-Sp  & 70.7  &  60.3  &  84.0  &  67.8  &  70.7 \\
L32-3B-IT-It & 74.9  &  56.6  &  83.6  &  66.2  &  70.3 \\
L32-3B-IT-Ko & 69.6  &  58.8  &  80.9  &  60.1  &  67.3 \\
L32-3B-IT-Ch & 69.3  &  58.3  &  79.7  &  59.3  &  66.7 \\
\midrule
Q25-3B-IT-En & 70.7  &  61.6  &  85.4  &  73.6  &  72.8 \\
Q25-3B-IT-Sp & 74.9  &  59.6  &  82.3  &  69.2  &  71.5 \\
Q25-3B-IT-It & 74.3  &  62.1  &  82.0  &  69.4  &  71.9\\
Q25-3B-IT-Ko & 68.4  &  63.2  &  80.9  &  61.4  &  68.5\\
Q25-3B-IT-Ch & 74.3  &  61.2  &  82.2  &  66.2  &  71.0\\
\midrule
Q25-7B-IT-En & 68.2  &  66.2  &  87.9  &  70.9  &  73.3  \\
Q25-7B-IT-Sp & 75.7  &  59.9  &  86.1  &  70.4  &  73.0  \\
Q25-7B-IT-It & 76.3  &  61.0  &  84.9  &  68.8  &  72.7   \\
Q25-7B-IT-Ko & 72.9  &  65.4  &  84.8  &  67.6  &  72.7   \\
Q25-7B-IT-Ch & 76.3  &  63.2  &  84.6  &  65.1  &  72.3   \\
\bottomrule
\end{tabular}
\end{sc}
\end{small}
\end{center}
\caption{Korean \textsc{RewardBench} results for reward model comparison across four different categories. (* denotes off-the-shelf models)}
\label{table:rb_kor}
\vskip -0.1in
\end{table*}

\begin{table*}[ht!]
\vskip 0.15in
\begin{center}
\begin{small}
\begin{sc}
\begin{tabular}{l|c|c|c|c|c}
\toprule
\textbf{Reward Model} & \textbf{Chat} & \textbf{Chat(H)} & \textbf{Safety} & \textbf{Reason} & \textbf{Avg.} \\
\toprule
ArmoRM-L3-8B* & 60.6  &  60.5  &  83.7  &  73.6  &  69.6  \\ 
\midrule

L32-3B-IT-En & 54.7  &  64.0  &  82.6  &  79.3  &  70.2 \\
L32-3B-IT-Sp & 61.2  &  60.5  &  82.9  &  70.5  &  68.8  \\
L32-3B-IT-It & 66.8  &  57.0  &  84.9  &  66.4  &  68.8  \\
L32-3B-IT-Ko & 68.4  &  61.0  &  81.1  &  61.3  &  67.9\\
L32-3B-IT-Ch & 68.7  &  59.9  &  81.2  &  52.6  &  65.6\\
\midrule
Q25-3B-IT-En & 58.7  &  67.8  &  84.3  &  78.2  &  72.2 \\
Q25-3B-IT-Sp & 68.7  &  62.5  &  79.5  &  71.0  &  70.4 \\
Q25-3B-IT-It & 69.8  &  62.3  &  81.6  &  70.6  &  71.1 \\
Q25-3B-IT-Ko & 70.1  &  61.4  &  79.7  &  62.3  &  68.4 \\
Q25-3B-IT-Ch & 69.8  &  64.7  &  81.8  &  61.3  &  69.4 \\
\midrule
Q25-7B-IT-En & 55.0  &  66.2  &  85.7  &  75.8  &  70.7  \\
Q25-7B-IT-Sp & 71.5  &  63.4  &  84.9  &  72.9  &  73.2   \\
Q25-7B-IT-It & 70.9  &  60.7  &  85.7  &  67.6  &  71.2  \\
Q25-7B-IT-Ko & 73.5  &  60.7  &  83.9  &  70.1  &  72.1  \\
Q25-7B-IT-Ch & 67.9  &  61.6  &  84.8  &  64.1  &  69.6   \\
\bottomrule
\end{tabular}
\end{sc}
\end{small}
\end{center}
\caption{Chinese \textsc{RewardBench} results for reward model comparison across four different categories. (* denotes off-the-shelf models)}
\label{table:rb_ch}
\vskip -0.1in
\end{table*}

\clearpage

\end{document}